# A Review of Intelligent Device Fault Diagnosis Technologies Based on Machine Vision

Guiran Liu*
San Francisco State University, College of Science & Engineering (CoSE)
San Francisco, United States
gliu@sfsu.edu

Binrong Zhu
San Francisco State University, College of Science & Engineering (CoSE)
San Francisco, United States
bzhu2@sfsu.edu

*Abstract-*: This paper provides a comprehensive review of mechanical equipment fault diagnosis methods, focusing on the advancements brought by Transformer-based models. It details the structure, working principles, and benefits of Transformers, particularly their self-attention mechanism and parallel computation capabilities, which have propelled their widespread application in natural language processing and computer vision. The discussion highlights key Transformer model variants, such as Vision Transformers (ViT) and their extensions, which leverage self-attention to improve accuracy and efficiency in visual tasks. Furthermore, the paper examines the application of Transformer-based approaches in intelligent fault diagnosis for mechanical systems, showcasing their superior ability to extract and recognize patterns from complex sensor data for precise fault identification. Despite these advancements, challenges remain, including the reliance on extensive labeled datasets, significant computational demands, and difficulties in deploying models on resource-limited devices. To address these limitations, the paper proposes future research directions, such as developing lightweight Transformer architectures, integrating multimodal data sources, and enhancing adaptability to diverse operational conditions. These efforts aim to further expand the application of Transformer-based methods in mechanical fault diagnosis, making them more robust, efficient, and suitable for real-world industrial environments.

*Keywords- Transformer Networks; Fault Diagnosis；Vibration Analysis；Deep Learning；*

## I. Introduction

With the continuous advancement of modern technology, mechanical equipment used in industrial applications has become increasingly systematized, automated, and intelligent, with more diverse and complex functional structures[1]. These machines are widely used across industries such as aerospace, transportation, power generation, automotive manufacturing, and mechanical processing, including applications like aircraft engines, wind turbines, industrial gearboxes, high-speed trains, and construction machinery[2]. As industrial production demands higher speed, load capacity, and automation, equipment failures can lead to significant downtime, causing substantial economic losses and even potential casualties[3]. It is estimated that mechanical equipment failures contribute to approximately 38%[4] of major accidents and economic losses in industrial production. Therefore, real-time monitoring and fault diagnosis have become critical for ensuring smooth operation and preventing serious accidents[5].

The operation of mechanical equipment is often accompanied by complex physical phenomena such as vibration, acoustic radiation, and heat transfer, which contain vital information about the equipment's condition. Vibration signals, in particular, can reveal early-stage faults, making them a key focus in fault diagnosis research[6]. Consequently, fault diagnosis plays a crucial role in system design and maintenance, enhancing economic efficiency. However, modern mechanical equipment often displays characteristics like fault coupling, delays, and hierarchical failure patterns, which present challenges for effective diagnosis and call for more advanced methods[7].

The process of fault diagnosis typically includes signal acquisition, feature extraction, and pattern recognition. Traditional approaches rely on sensor data visualization and predefined thresholds (such as temperature, vibration, or speed) to monitor equipment health. These methods, however, struggle to identify early faults accurately and promptly. With the rapid rise of artificial intelligence (AI), intelligent fault diagnosis methods have gained significant attention, particularly those based on machine learning (ML)[8] and, more recently, deep learning (DL). Deep learning models, such as convolutional neural networks (CNN) and recurrent neural networks (RNN), have proven effective by automatically extracting deep features from data, reducing the need for manual intervention and improving fault detection accuracy[9].

In recent years, Transformer-based methods have shown great potential for intelligent fault diagnosis in mechanical equipment, benefiting from their powerful feature extraction and pattern recognition capabilities. While there have been several studies on this topic, the research is scattered across various sources and lacks comprehensive review. Therefore, conducting a systematic and thorough analysis of Transformer-based fault diagnosis methods is crucial for understanding the current state of this emerging technology and its future directions[10]. Such an effort would raise awareness in both academic and industrial circles, facilitating the development and application of Transformer-based solutions in the field of mechanical equipment fault diagnosis.

## II. The Progress and Limitations of Existing Fault Diagnosis Methods for Mechanical Equipment

Mechanical equipment fault diagnosis involves analyzing performance data to identify specific fault types, traditionally using either physical model-based methods or AI-based approaches. Physical model methods focus on the evolution of failure mechanisms like wear, cracks, and fatigue, but building accurate models requires expert knowledge and assumptions, making it challenging for complex systems. AI-based methods, on the other hand, have gained popularity due to advances in computer technology and data science, offering simplicity, broad applicability, and independence from detailed physical models. These AI models use intelligent algorithms to process sensor data, extracting features that accurately represent equipment conditions and identifying fault patterns. While earlier AI methods relied on traditional machine learning, modern deep learning techniques have demonstrated superior abilities in feature extraction and intelligent decision-making, opening new possibilities for effective fault diagnosis. Signal acquisition and feature extraction are critical stages where understanding physical principles, phenomena, and models is essential. Physical principles like wave motion and resonance provide the theoretical basis for signal analysis, while physical phenomena, such as changes in amplitude or frequency, indicate different fault types, offering strong diagnostic evidence. Moreover, physical models act as a bridge, abstracting real systems into simplified representations that simulate operational states and fault processes, generating valuable training data for AI models. By integrating both approaches, this review outlines the current progress and challenges in mechanical fault diagnosis, pointing to future opportunities for more efficient and intelligent diagnostic systems.

### A. Signal acquisition

Currently, the common types of signals collected during the monitoring of mechanical equipment include vibration signals[11], sound signals, temperature signals, as well as data from oil analysis instruments and infrared imaging. Each of these methods exhibits varying sensitivity and maintenance costs in the context of mechanical fault diagnosis.

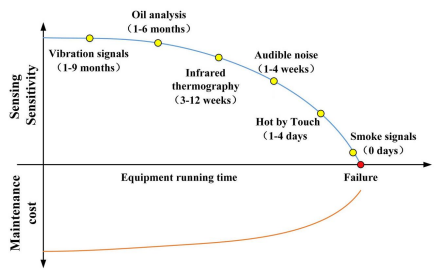

Figure 1: Sensitivity and associated maintenance costs of different sensor technologies in mechanical fault diagnosis.

As shown in Figure 1, it is evident that high sensor sensitivity enables the detection of certain equipment anomalies through signal analysis. However, in practical applications of mechanical equipment, this sensitivity may hinder accurate fault localization and resolution, resulting in lower diagnostic accuracy and increased maintenance costs. While advancements in sensor technology enhance the ability to identify faulty components and fault patterns, thereby reducing maintenance expenses, they also raise the likelihood of equipment failures.

### B. Feature extraction

Feature extraction is a crucial step in the fault diagnosis process, as it aids in identifying patterns and structures within data, thereby providing better inputs for determining fault types. The quality of feature extraction directly influences the effectiveness of fault diagnosis, leading to extensive research and practical exploration in this domain. Common feature extraction methods include time-domain features, which capture the characteristics and statistics of signals over time, such as mean, peak value, root mean square (RMS)[12], and entropy. Frequency-domain features, on the other hand, characterize the properties of signals in the frequency domain, offering insights into frequency content and distribution through parameters like center frequency and frequency variance. Additionally, time-frequency analysis methods, such as Short-Time Fourier Transform and Wavelet Transform, have been integrated into mechanical fault diagnosis to address the non-smooth and nonlinear nature of vibration and acoustic signals. These methods decompose signals into different components, creating two-dimensional representations that facilitate pattern recognition. Furthermore, advancements in computer vision have led to image-based feature extraction techniques, including color, texture, shape, and deep learning-based methods, which leverage neural networks to learn high-level feature representations from images. Lastly, text features extracted from textual data, such as text length, syntactic structures, and sentiment analysis, further contribute to the comprehensive understanding of mechanical systems for effective fault diagnosis.

## III. The application of transformers in mechanical equipment fault diagnosis

### A. The network architecture and principles of the Transformer

The Transformer model, designed for sequence-to-sequence tasks, leverages the self-attention mechanism as its foundation. Before the advent of the Transformer, RNNs dominated as the most commonly applied models in the field of NLP[13], with their structure shown in Figure 2.

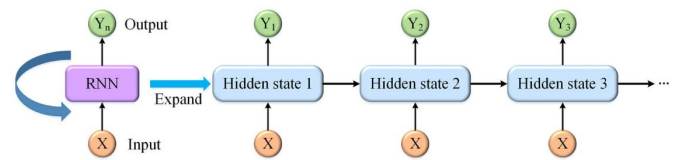

Figure 2.The standard architecture of RNN.

The Transformer model was initially utilized in natural language processing for machine translation tasks, achieving significant results. In recent years, it has been creatively applied in the computer vision domain, contributing to image enhancement, generation, classification, object detection, and segmentation, thus creating new milestones in the field. The Transformer consists of three main components: the encoder, decoder, and positional encoding. The encoder generates input encodings, while the decoder receives these encodings to merge contextual information and produce the output sequence. Each module of the Transformer is described in detail[14].

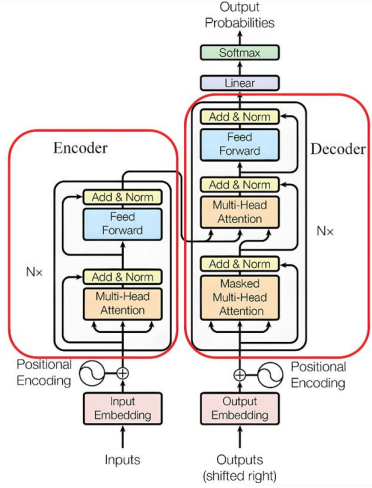

Figure 2: Original Structure of the Transformer

The Transformer employs an encoder-decoder model architecture that eliminates recurrence, as illustrated in Figure 6. The first component is the encoder, which consists of six identical stacked encoder layers. Each encoder layer comprises two sub-layers: Multi-Head Self-Attention (MSA) and Feed-Forward Neural Network (FFN)[15]. The MSA mechanism enables the model to focus on different positions within the input sequence, capturing global contextual information. The FFN is used for applying non-linear transformations to the features at each position. By stacking multiple encoder layers, the encoder progressively extracts an abstract representation of the input sequence.

FFN is a fully connected feedforward neural network added after the self-attention layers in both the encoder and decoder. It takes the output from the self-attention layer as input and produces a new representation vector that encapsulates more advanced semantic information. The computation process of FFN is as follows:

$$\text{FFN}(X) = W_2 \sigma(W_1 X) \quad (1)$$

In the FFN, W1 and W2 represent the linear transformation matrices of the first and second fully connected layers, respectively, while p denotes the nonlinear activation function. The dimension of the hidden layer is dh=2,048d_h = 2,048dh =2,048.

The FFN utilizes a two-layer fully connected structure, with a ReLU activation function applied between the layers. Specifically, in each FFN, the input representation vector first undergoes a linear transformation through a fully connected layer, then a nonlinear transformation via the ReLU activation function, and finally, another linear transformation through the second fully connected layer to produce the output. The advantage of FFN lies in its ability to extract higher-level semantic features from the input through multiple layers of nonlinear transformations, enhancing the model's expressive power. Additionally, since the computation in the FFN is independent, it can be parallelized, thereby accelerating the model's training process.

Since the Transformer does not include any recurrent or convolutional structures to capture the positional information of words in a text, it is necessary to incorporate some relative or absolute position information of tokens in the sequence, allowing the model to capture the relationships between sequential data. To achieve this, position encodings are added at the bottom of both the encoder and decoder stacks[16]. Each word in the text is assigned a position number, which corresponds to a word vector, and the position vector is combined with the word vector, embedding the positional information into each word.

Compared to the sequential input method of RNNs, the Transformer allows for parallel data input while maintaining the positional relationships between data, which enhances computational speed and reduces storage requirements. Additionally, the dimensions of the position encoding and the input sequence embedding vectors are the same, allowing them to be added together. Currently, there are various methods for position encoding, and the Transformer uses sinusoidal functions with different frequencies to encode position information, preserving the relative relationships between positions. The specific computation can be expressed as:

$$\text{PE}_{(\text{pos}, 2i)} = \sin(\text{pos}/10{,}000^{2i/d_m}) \quad (1)$$

$$\text{PE}_{(\text{pos}, 2i+1)} = \cos(\text{pos}/10{,}000^{2i/d_m}) \quad (1)$$

Here, pos represents the position of each word in the text, iii denotes the dimension, and dm refers to the dimension of the position encoding. The term 2i represents the even dimensions of the position encoding, while 2i+12i+12i+1 corresponds to the odd dimensions of the position encoding (where 2i≤d2, 2i+1≤d).

As can be seen, each dimension of the position encoding corresponds to a sine wave with wavelengths ranging from $2\pi$ to $10{,}000^{-2\pi}$ in a geometric progression.

The attention mechanism, inspired by neuroscience, allows models to dynamically allocate varying attention weights to different parts of the input data, enabling selective processing of critical information while disregarding irrelevant details. This capability enhances the model's ability to understand input data and extract key features, making the mechanism particularly useful in domains such as speech recognition, machine translation, and image processing. Self-attention further refines this mechanism by focusing on the internal relationships between different parts of the input data, whereas traditional attention mechanisms emphasize correlations between elements within sequences. Unlike traditional methods, self-attention operates independently of external information, relying solely on the intrinsic characteristics of the input data. It excels at capturing long-range dependencies, overcoming limitations often faced by conventional attention mechanisms in such scenarios. By fully embracing self-attention for global feature extraction, Transformers eliminate the need for convolutional and recurrent operations. The original Transformer introduced scaled dot-product attention (SDPA)[17], whose structure is illustrated in Figure 3.

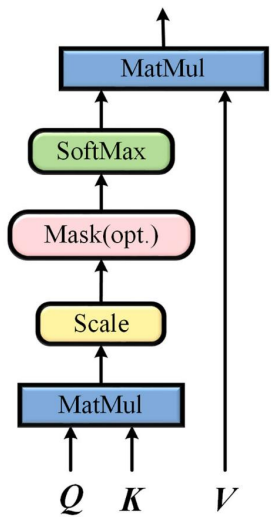

Figure 3.The fundamental structure of scaled dot-product attention (SDPA).

### B. Transformer-based image classification model

In the field of computer vision, the original Transformer model is not widely used, as it was primarily designed for sequential data processing tasks. However, ongoing research has led to improvements that allow its application in image processing, yielding significant success. While Transformers excel in handling sequential data like text, image data can also be viewed as a two-dimensional sequence. This insight has inspired researchers to adapt Transformer models for image tasks, resulting in several high-performing visual Transformer models. Among these, image classification is a prominent application, where the goal is to differentiate images based on their embedded meanings and contextual information, serving as a foundation for other image processing activities such as object detection and image segmentation. To enhance the efficiency of visual Transformer models, researchers have introduced various modifications to the original architecture. This study focuses on several high-performing visual Transformer models, particularly ViT and its variants, summarizing their research advancements.

ViT represents the first successful application of the Transformer architecture in image classification, surpassing the state-of-the-art CNN models of its time, such as ResNet and EfficientNet, in both classification performance and model architecture, as demonstrated in Figure 4. This advancement marked a significant milestone in the use of Transformers for visual tasks.

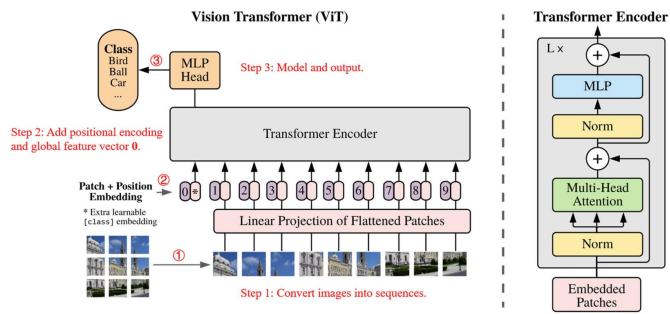

Figure 4: Structure of the ViT Model

The key to using ViT for image classification lies in transforming the image into a sequence of data[18]. ViT segments the input image into a series of patches, with each patch containing a portion of the image's information. Subsequently, each patch is converted into a vector representation, known as an embedding.

By leveraging knowledge distillation and self-supervised learning, DeiT achieves image classification results on the ImageNet dataset comparable to top-tier CNN models while requiring less data and computational resources. The structural design of the DeiT model is illustrated in Figure 5[19].

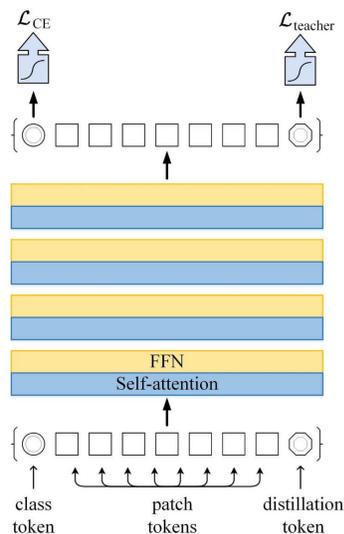

Figure 5 depicts the structural architecture of the DeiT model.

As illustrated in Figure 5, the training process of DeiT consists of four main stages: data preparation, model architecture, self-supervised pretraining, and supervised fine-tuning. Compared to traditional ViT, DeiT achieves superior results with fewer samples or even without any labeled samples.

### C. Methods for Intelligent Fault Diagnosis of Mechanical Equipment Based on Transformers

The original Transformer model is primarily designed for natural language processing and is not directly applicable to image data. However, recent modifications have enabled its use in visual tasks like image recognition. Two main approaches for applying Transformer-based methods in mechanical equipment fault diagnosis emerge from the literature. The first involves preprocessing one-dimensional fault signals, such as vibrations and sounds, to convert them into a format suitable for Transformer input, allowing for effective feature extraction. These signals can be analyzed through their characteristics like frequency and amplitude. The second approach transforms these one-dimensional signals into two-dimensional images using time-frequency methods, which are then input into visual Transformer models like ViT or Swin Transformer for training and fault pattern recognition. Datasets used for validating these methods include the Case Western Reserve University (CWRU)[20] bearing dataset and others, which contain various fault modes and lifecycle vibration data, with data collection arrangements illustrated in relevant figures 6.

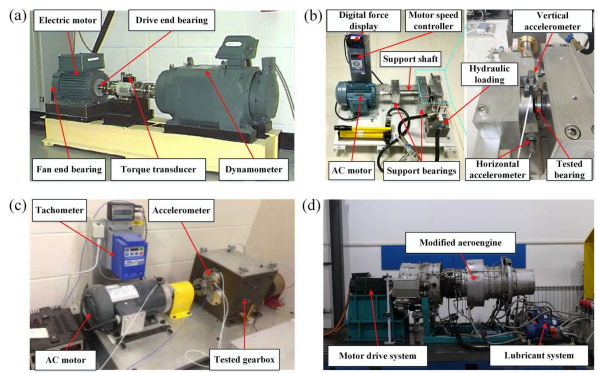

Figure 6 Data collection equipment: (a) CWRU dataset; (b) XJTU-SY dataset; (c) UCONN dataset; (d) HIT dataset.

The CWRU bearing dataset includes 10 predefined rolling bearing faults, with a normal condition as a special mode, making it vital for rotating machinery fault diagnosis. The XJTU-SY dataset documents the full life cycle vibration data of 15 rolling bearings across three operating conditions, while the UCONN dataset contains 936 samples with nine failure modes from two-stage gearboxes. The HIT dataset, based on a real aviation engine, features three states of inter-shaft bearings.

These datasets illustrate key physical phenomena. Normal operation produces periodic vibration signals with distinct frequency peaks, while faults create transient impacts that serve as important diagnostic indicators. Signal amplitudes and frequency components evolve with performance decline, often increasing or changing. Resonance occurs when vibration frequencies approach inherent frequencies, amplifying vibrations and aiding fault diagnosis. Energy transfer among components can also reveal abnormalities, and the vibration signals can be analyzed using wave principles to support effective fault diagnosis.

Table 1 Performance of various methods across four public datasets

| Methods | Classification accuracy (%) |
|---|---|
| **CWRU dataset** | |
| BPNN | 81.35 |
| DBN | 88.20 |
| 1DCNN | 97.32 |
| TST | 98.63 |
| Diagnosisformer | 99.85 |
| SiT | 99.46 |
| TAR | 99.90 |
| SViT | 97.56 |
| **XJTU-SY dataset** | |
| NKH-KELM | 95.56 |
| DCN | 99.31 |
| AlexNet | 99.58 |
| LSTM | 98.65 |
| CWT-2DCNN | 99.40 |
| TST | 99.78 |
| **UCONN dataset** | |
| AE | 95.13 |
| DAE | 93.76 |
| BPNN | 95.13 |
| LSTM | 88.74 |
| ResNet18 | 85.84 |
| TST | 99.51 |
| **HIT dataset** | |
| CNN | 83.13 |
| LSTM | 85.41 |
| TST | 71.07 |

From the analysis of Table 1, it is evident that the fault diagnosis accuracy on the HIT dataset is significantly lower than that of the other three datasets. This is because the HIT dataset closely reflects real-world mechanical equipment fault diagnosis scenarios, making it more representative of practical conditions. Unlike datasets collected under controlled laboratory settings, the HIT dataset presents greater challenges, establishing a new benchmark for validating mechanical fault diagnosis methods.

Driven by the pioneering research concept of transformer-based intelligent fault diagnosis for mechanical equipment, Jin et al. introduced a Time-Series Transformer (TST)-based approach for diagnosing faults in rotating machinery. This method addresses the long-term dependency issues inherent in traditional CNN- and RNN-based fault diagnosis models[21]. The overall architecture of the TST model is depicted in Figure 7.

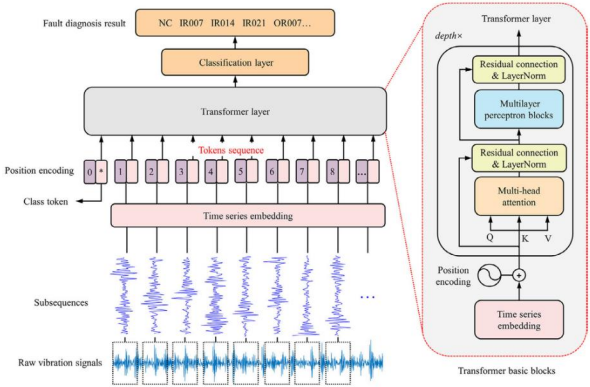

Figure 7 illustrates the comprehensive architecture of the TST model.

As illustrated in Figure 7, a novel time series annotator was designed specifically for one-dimensional data processing, forming the basis for the TST model that integrates the Transformer architecture. The model's effectiveness was validated using the CWRU dataset, XJTU-SY dataset, and UCONN dataset. Experimental results demonstrated that TST achieved fault diagnosis accuracies of 98.63% (10 classes), 99.72% (4 classes), 99.78%, and 99.51%, respectively, outperforming traditional CNN and RNN models. Furthermore, feature visualization using t-SNE revealed that the feature vectors extracted by TST exhibit superior intra-class compactness and inter-class separability, further substantiating the method's effectiveness[21].

Hou et al. [22] addressed common challenges in DL-based bearing fault diagnosis models driven by big data, such as difficulties in data acquisition, imbalanced class distributions, and noise interference. They proposed a fault diagnosis approach based on Transformer and ResNet joint feature extraction (TAR). The overall architecture of the TAR model is shown in Figure 8.

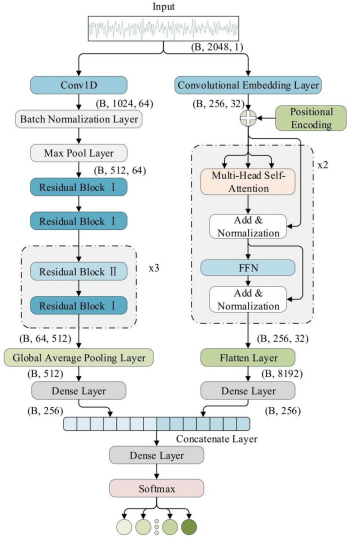

Figure 8 presents the overall architecture of the TAR model.

As depicted in Figure 8, the raw 1D signal is initially processed by a 1D convolutional layer for feature separation and embedding. These features are then passed into the Transformer encoder and ResNet framework for further extraction. This approach demonstrates superior diagnostic accuracy compared to traditional deep learning networks. Additionally, a transfer learning strategy with model fine-tuning helps reduce the training difficulty for new tasks. The model's effectiveness was validated using the CWRU dataset, with experimental results showing that TAR achieves a fault diagnosis accuracy of 99.90% on the CWRU dataset without noise. Even with varying levels of noise added, TAR[23] consistently outperforms the comparison methods in terms of average fault diagnosis accuracy.

The raw 1D input signal is first normalized within the range of [-1, 1], and a convolutional embedding module is introduced to replace the original embedding module. To reduce model complexity, linear self-attention is used in place of the traditional self-attention, ensuring that CLFormer meets the lightweight requirements. The proposed method's effectiveness was evaluated on a rotary machinery dataset collected in a laboratory. The experimental results demonstrate that, compared to the Transformer, CLFormer reduces the number of parameters from 35.22K to 4.88K and improves fault diagnosis accuracy from 82.68% to 90.53%, indicating its practical application potential.

To solve the problems of low accuracy and poor robustness in traditional deep learning-based rolling bearing fault diagnosis, Hou et al. designed a multi-feature parallel fusion rolling bearing fault diagnosis method based on the Transformer network, called Diagnosisformer. The overall architecture of Diagnosisformer is shown in Figure 9.

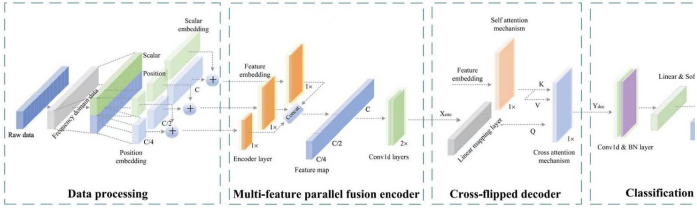

Figure 9 illustrates the overall model architecture of Diagnosisformer.

The Fast Fourier Transform (FFT) is primarily employed to extract frequency-domain features from raw one-dimensional vibration data. These features are normalized and embedded into the network as inputs. The multi-feature parallel fusion encoder is then utilized to capture both local and global features of the bearing data. These extracted features are passed to the cross-reverse decoder and subsequently classified by the classification head for fault diagnosis. The proposed model's effectiveness was validated using self-constructed rotating machinery fault diagnosis data and the CWRU dataset. Experimental results demonstrated that Diagnosisformer achieved average diagnostic accuracies of 99.84% and 99.85% on the two datasets, respectively, significantly outperforming methods such as CNN, CNN-LSTM, RNN, LSTM, and GRU in terms of both accuracy and robustness[24].

Yang et al. introduced a Signal Transformer (SiT) based on attention mechanisms and applied it to bearing fault diagnosis research. The overall model architecture of SiT is shown in Figure 10.

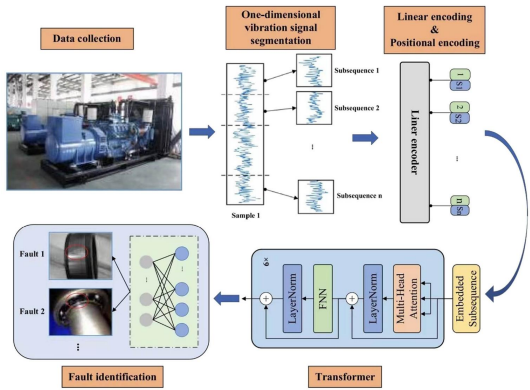

Figure 10 illustrates the overall architecture of the Signal Transformer (SiT).

Figure 10 illustrates the workflow, which starts by dividing the original one-dimensional vibration time series into multiple subsequences. These subsequences undergo both linear encoding and positional encoding before being processed by the Transformer to extract features for identifying fault patterns in bearings. The proposed method was tested on the CWRU bearing dataset and a self-constructed centrifugal pump bearing dataset. The experimental outcomes revealed that SiT achieved average diagnostic accuracies of 99.46% and 99.53% for the respective datasets, effectively demonstrating the approach's validity[25].

### D. Overview of Current Research Status

Current research indicates that significant advancements have been made in applying Transformer-based methods for intelligent fault diagnosis of mechanical equipment. The architecture's self-attention mechanism, parallel processing capabilities, and flexible structure have attracted considerable interest. Since its introduction, the Transformer has gained rapid traction, achieving impressive results in natural language processing, speech recognition, and image processing. However, its application in mechanical fault diagnosis remains limited, facing challenges such as complex architecture requiring large datasets, which increases the risk of overfitting and reduces effectiveness in small datasets or specialized tasks. Most studies focus on stable operating conditions and analyze only single fault modes, which is unrealistic given that machinery often experiences complex faults from multiple sources. While Transformers are commonly used for rotating machinery like bearings and rotor systems, research on reciprocating machinery, such as diesel engines, is sparse. Despite these challenges, Transformer-based methods have demonstrated remarkable effectiveness and accuracy in fault diagnosis, revealing strong correlations between specific vibration signals and mechanical faults. These findings not only offer new diagnostic indicators but also enhance understanding of the operational mechanisms and failure processes in mechanical systems.

## IV. Research prospect

With advancements in fault diagnosis methods and technologies for machinery, Transformer-based intelligent fault detection has become a rising and significant area of research. In recent years, Transformers have gained widespread application across various artificial intelligence domains, such as computer vision (CV), natural language processing (NLP), and multimodal analysis, establishing themselves as a leading methodology. As deep learning (DL) technologies continue to evolve, the application of Transformer-based diagnostic approaches is anticipated to expand further. This discussion highlights current challenges and outlines the future development directions for Transformer-based methods in machinery fault diagnosis.

**Enhanced Adaptability to Diverse Data**

Currently, the application of Transformer-based diagnostic techniques predominantly focuses on rotating machinery, including rolling bearings and rotor systems, using data such as vibration and sound signals. However, with the emergence of the Industrial Internet, operational data types like temperature, pressure, and flow rates are expected to be integrated into machinery fault diagnosis. This integration will expand the applicability of these techniques, enabling their use in diagnosing reciprocating machinery such as engines. The inclusion of diverse data sources and types will also enhance the reliability and accuracy of diagnostic outcomes.

**Addressing Data Scarcity with Few-Shot Learning**

Diagnostic models built on Transformers often rely on extensive datasets for training, followed by fine-tuning to achieve task-specific accuracy and generalizability. However, in many fault diagnosis scenarios, gathering large volumes of labeled data is challenging due to the rarity and randomness of machinery faults. Few-shot learning presents a potential solution by utilizing algorithms and model enhancements to extrapolate knowledge from a limited number of samples, enabling predictions on unseen data. This approach effectively addresses the Transformer's limitations in scenarios where small, specialized datasets are available.

**Development of Lightweight Transformers**

Although Transformers deliver exceptional performance in fault diagnosis tasks, their high computational complexity and substantial parameter requirements can hinder deployment in real-world applications. Specifically, in resource-constrained environments such as embedded systems or mobile devices, the need for extensive computational resources and storage can pose significant challenges. Research efforts focusing on lightweight Transformers aim to simplify models through pruning, compression, and optimization techniques. These improvements reduce memory requirements, accelerate inference speeds, and enhance performance in scenarios demanding real-time diagnostics, such as speech processing or image recognition in mechanical systems.

**Synergy Between Transformers and CNNs**

Transformers are particularly effective in processing long-sequence data and capturing global dependencies but may underperform with short-sequence data compared to CNNs. Conversely, CNNs excel at extracting local features, particularly for image and text data, but lack the ability to effectively capture global patterns. Combining the strengths of Transformers and CNNs allows for models that can simultaneously learn global and local features. This integration maximizes representational and generalization capabilities, creating robust diagnostic systems adaptable to a variety of tasks and datasets.

By addressing these future directions, Transformer-based approaches for machinery fault diagnosis can become more versatile, efficient, and robust, enabling broader and more impactful applications in industrial settings.

*V. Conclusion*

This study underscores the significant potential of Transformer-based models for intelligent fault diagnosis in mechanical systems. These models have demonstrated exceptional performance in capturing complex fault patterns and achieving high diagnostic accuracy, particularly in controlled scenarios. However, several challenges persist, including the high computational complexity of Transformer architectures and their reliance on extensive labeled datasets for training. Additionally, current research predominantly emphasizes stable operational conditions and isolated fault modes, which may not adequately reflect the diverse and multifaceted nature of real-world machinery faults.

Moreover, there remains a gap in applying Transformer-based methods to diverse industrial contexts, particularly in reciprocating machinery like diesel engines and other equipment beyond rotating machinery. The findings of this study highlight the ability of Transformers to leverage frequency-domain features from vibration signals, indicating their strong capability to extract and learn fault-relevant patterns. However, their practical implementation in resource-constrained or highly dynamic industrial settings remains limited.

To address these gaps, future research should focus on developing lightweight and adaptable Transformer architectures, integrating diverse operational data types, and enhancing the models' robustness under varying and unpredictable conditions. Furthermore, exploring hybrid approaches, such as combining Transformers with other deep learning techniques like CNNs, can balance computational efficiency and diagnostic precision. By overcoming these challenges, Transformer-based methods can evolve into versatile and scalable solutions, expanding their applicability in industrial fault diagnosis and contributing significantly to predictive maintenance and operational efficiency.